\documentclass[preprint,12pt]{elsarticle}

\usepackage[english]{babel}

\usepackage{subfigure}

\usepackage{subfigure}
\usepackage{graphicx}
\usepackage{epsfig}

\usepackage{hyperref}

\hypersetup{
    colorlinks,%
    citecolor=black,%
    filecolor=black,%
    linkcolor=black,%
    urlcolor=black
}

\usepackage{amssymb}

\journal{Atmospheric Enviroment}

\begin{document}

\begin{frontmatter}

\title{Neural-estimator for the surface emission rate of atmospheric gases}

\author[rvt,inst]{F.F.˜Paes}
\ead{fabiana.paes@lac.inpe.br}
\author[focal,inst]{H.F.˜Campos Velho}
\ead{haroldo@lac.inpe.br}

\address[rvt]{Departamento de Computa\c c\~ao Aplicada}
\address[focal]{Laborat\'orio Associado de Computa\c c\~ao e Matem\'atica Aplicada}
\address[inst]{Instituto Nacional de Pesquisas Espaciais, box 515, S\~ao Jos\'e dos Campos-SP-Brazil}

\begin{abstract}
The emission rate of minority atmospheric gases is inferred by a new approach based on neural networks. The new network applied is the multi-layer perceptron with backpropagation algorithm for learning. The identification of these surface fluxes is an inverse problem.  A comparison between the new neural-inversion and regularized inverse solutions is performed. The results obtained from the neural networks are significantly better. In addition, the inversion with the neural networks is faster than regularized approaches, after training.

\end{abstract}


\end{frontmatter}


\section{Introduction}
\label{introduction}


The enhancing of the concentration of greenhouse effect gases is a central issue nowadays, meanly regarding the most important anthropogenic gases, such as methane ($CH_{4}$) and carbon dioxide ($CO_{2}$). Despite the ratification of the Kyoto Protocol, the forecast is that the releases of $CO_{2}$ and $CH_{4}$ in the atmosphere continue to increase in next decade \cite{ipcc}.

One mandatory strategy is to monitoring the concentration of these gases in the atmosphere. However, in order to understand the bio-geochemical cycle of these gases, it is necessary to estimate the surface emission rates. One procedure for this is to employ inverse problem methodology.

The method of inverse problem is an efficient way to scientifically estimate the intensity of pollution sources. Various inverse problem methods are being investigated by the international scientific community \cite{enting, seibert, seibert2001}. In order to deal with the ill-posed characteristic of inverse problems, regularized solutions \cite{camposvelho97, tikhonov} and also regularized iterative solutions \cite{alifanov, chiwiacowsky} have been proposed. More recently, artificial neural networks are also employed to solve inverse problems \cite{hidalgo, woodbury, shiguemori}. The pollutant source identification is an inverse problem, and neural networks have been applied for identifying the emission intensity of point sources \cite{kukkonen, lu, gardner, pelliccioni, wesolowski}.

In this paper, a new approach using multilayer perceptron artificial neural network (MLP-ANN) is employed to estimate the rate of surface emission of a pollutant. The input for the ANN is the gas concentration measured on a set of points. The methodology is tested using synthetic experimental data, obtained by running an atmospheric pollutant dispersion model: LAMBDA \cite{ferrero95, ferrero1998a}. The Lambda model is a Lagrangian model. 

Finally, the surface rate estimated with MLP-ANN is compared with regularized inversion by maximum entropy principle. In the latter method, the inverse problem is formulated as an optimization problem that could be solved using a deterministic or a stochastic optimization procedure. 

\section{Forward Model}
\label{fm}

The Lagrangian particle model LAMBDA was developed to study the transport process and pollutants diffusion, starting from the Brownian random walk modeling \cite{ferrero1998a, ferrero1998b}. In the LAMBDA code, full-uncoupled particle movements are assumed. Therefore, each particle trajectory can be described by the generalized three dimensional form of the Langevin equation for velocity \cite{thomson}:

\begin{eqnarray} \label{um}    
du_{i}=a_{i}\left(\textbf{x},\textbf{u},t\right)dt+b_{ij}\left(\textbf{x},\textbf{u},t\right)dW_{j}(t)
\end{eqnarray}

\begin{eqnarray} \label{dois}    
d\textbf{x}=\left(\textbf{U}+\textbf{u}\right)dt
\end{eqnarray}
where $i,j=1,2,3$, and \textbf{x} is the displacement vector, \textbf{U} is the mean wind velocity vector,\textbf{u} is the Lagrangian velocity vector, $a_{i}\left(\textbf{x},\textbf{u},t\right)$ is a deterministic term and $b_{ij}\left(\textbf{x},\textbf{u},t\right)dW_{j}(t)$ is a stochastic term and the quantity $dW_{j}(t)$ is the incremental Wiener process.

The determinisitc (drift) coefficient $a_{i}\left(\textbf{x},\textbf{u},t\right)$ is computed using a particular solution of the Fokker-Planck equation associated to the Langevin equation. The diffusion coefficient $b_{ij}\left(\textbf{x},\textbf{u},t\right)$ is obtained from the Lagrangian structure function in the inertial subrange $\left(\tau_{K}<<\Delta t<<\tau_{L}\right)$, where $\tau_{K}$ is the Kolmogorov time scale and $\tau_{L}$ is the Lagrangian de-correlation time scale. These parameters can be obtained employingthe Taylor statisitcal theory on turbulence \cite{degrazia2000}.

Backward integration can also be applied. This is just to identify which particle arriving in a sensor-$j$ is coming from a source-$i$.

The drift coefficient, $a_{i}\left(\stackrel{\rightarrow}{\textbf{x}},\stackrel{\rightarrow}{\textbf{u}},t\right)$, for forward and backward integration is given by

\begin{eqnarray} \label{tres}    
a_{i}P_{E}=c_{v}\left[\frac{\partial\left(B_{i,j}P_{E}\right)}{\partial\textbf{u}_{j}}+\phi_{i}\left(\textbf{x},\textbf{u},t\right)\right]
\end{eqnarray}
with

\begin{eqnarray} \label{quatro}    
\frac{\partial \phi_{i}}{\partial \textbf{u}_{i}}=-\frac{\partial P_{E}}{\partial t}-\frac{\partial \left(\textbf{u}_{i}P_{E}\right)}{\partial \textbf{x}_{i}}
\end{eqnarray}
and

\begin{eqnarray} \label{cinco}    
\phi_{i}\rightarrow0 \textit{when} \stackrel{\rightarrow}{\textbf{u}}\rightarrow \infty
\end{eqnarray}
where $c_{v}=1$ for forward integration and $c_{v}=-1$ for backward integration, $P_{E}=P\left(\stackrel{\rightarrow}{\textbf{x}},\stackrel{\rightarrow}{\textbf{u}},t\right)$ is the non-conditional PDF of the Eurelian celocity fluctuations, and $B_{i,j}=\frac{1}{2}b_{i,k}b_{j,k}$.

Of course, for backward integration, the time considered is: $t'=-t$, and velocity $\textbf{U}'=-\textbf{U}$, being \textbf{U} the mean wind speed. The horizontal PDFs are considered Gaussians, and for the vertical coordinate the truncated Gram-Charlier type-C of third order is employed \cite{anfossi}.

The diffusion coeffiecients, $b_{ij}\left(\textbf{x},\textbf{u},t\right)$, for both forward and backward integration is given by

\begin{eqnarray} \label{seis}    
b_{ij}=\delta_{ij}\left[2\frac{\sigma^{2}_{i}}{\tau_{Li}}\right]^{1/2}
\end{eqnarray}
where $\delta_{ij}$ is teh Kronecker delta, $\sigma^{2}_{i}$ and $\tau_{Li}$ are velocity variance at each component and the Lagrangian time scale \cite{degrazia2000}, respectively. With the coordinates and the mass of each particle, the concentration is computed - see equations (\ref{cinco}) and (\ref{seis}).

The inverse problem here is to identify the source term $S(t)$. As mentioned, a source-receptor approach is employed for reducing the computer time, instead of running the direct model (equation \ref{dois}) for each iteration. This approach displays an explicit relation between the pollutant concentration of the \textsl{i}-th receptor related the \textsl{j}-th sourcers:

\begin{eqnarray} \label{sete}    
C_{i}=\sum^{N_{S}}_{j=1}M_{ij}S_{j}
\end{eqnarray}
where the matrix $M_{ij}$ is the \textit{transition matrix}, and matrix entry given by

\begin{eqnarray} \label{oito}    
M_{ij}=\left\{\begin{array}{ll}\left(\frac{V_{S,j}}{V_{R,i}}\right)\left(\frac{\Delta t}{N_{S,j}}\right)N_{R,i,j} \left(\textit{forward}\right) \\ \left(\frac{\Delta t}{N_{S,j}}\right)N_{S,i,j} \left(\textit{backward}\right)
\end{array} \right.
\end{eqnarray}
where $V_R,i$ and $V_{S,j}$ are the volumr for the \textsl{i}-th receptor and \textsl{j}-th source, respectively; $N_{S,j}$ and $N_{R,i}$ are the number of particle realised by the \textsl{j}-th source and \textsl{i}-th sensor, respectively; $N_{R,i,j}$ and $N_{S,i,j}$ are the number of particle released by the \textsl{j}-th source and detected by the \textsl{i}-th receptor.

\section{Inverse Method: Neural Network}
\label{nn}

An artificial neural network (ANN) is an interconnected group of artificial neurons, elements of networks that uses a mathematical or computational model for information processing based on a connectionist approach to computation. Inputs and outputs to a neuron consist of values $\left(x_{1}, x_{2}, \ldots, x_{n}\right)$ and $\left(y_{1}, y_{2}, \ldots, y_{n}\right)$. The neuron computes the weighted sum of its inputs, adding a bias, and the result is an argument for a non-linear activation function. The MLP is composed of multiple processing units called artificial neurons (or nodes) arranged in several different layers \cite{bishop}. The configuration of the best MLP model includes choosing the number of layers (typically, it requires at least three: input, hidden, and output), the number of neurons in hidden layer (how many units should be in the input and output layers is defined by the problem), the activation function, and the learning algorithm. After the proper architecture of the MLP has been established, all the training cases are run through the network. In each neuron a linear combination of the weighted inputs (including a bias) is computed, summed and transformed using a transfer function (linear or nonlinear). The value obtained is passed on as an input to the neurons in the subsequent layer until a value is computed in neurons of the output layer. The output values are compared with the target values. The difference between the output and target is calculated for each output neuron using a certain error function in order to give the prediction error made by the network. Then, the training algorithm is used to adjust the network's weights and thresholds in order to minimize this error. Because a target value is compared to the output value, the learning process is called supervised \cite{haykin}. In this study a linear transfer function was used in input neurons, and the log-sigmoid function for the neurons located in hidden and output layers. The error function was the sum-squared error, where the individual errors of output units on each case were squared and summed together. The networks were trained using the back-propagation algorithm \cite{bishop}. The correctionof the weights $\Delta w$ was calculated according to the following formula: $\Delta w_{ij}=\eta\delta_{j}o_{i}+\alpha\Delta w^{\left(old\right)}_{ji}$, in wich \textit{j} is the index of the neuron in the current layer, the neuron in the upper layer is indexed by \textit{i} and its output by $o_{i}$, and the local error gradient is denoted by $\delta_{j}$. There are two constants in this formula - the learning rate $\eta$, wich determines to what extent the weights should be modified, and the momentum coefficient $\alpha$, which decides to what degree this previous adjustment is to be considered so as to prevent any sudden changes in the direction in which corrections are made. The learning rate $\eta$ and momentum $\alpha$ were set to 0.1 and 0.5, respectively.  They can be used to model complex relationships between inputs and outputs or to find patterns in data. In this paper, we used the neural network MLP feedforward backpropagation network is the multilayer perceptron (MLP) with backpropagation learning.

Regardless their type or use, all neural networks have three stages in their application: the learning, the activation and the generalization steps. It is in the learning step that the weights and bias corresponding to each connection are adjusted to some reference examples (the input). In the activation phase, the output is obtained based on the weights and bias computed in the learning phase \cite{haykin}. The transfer function selected was a tan-sigmoid transfer function for the hidden layers and linear transfer function for the output layer. The experimental data used here in the learning step were simulated adding a random perturbation to the exact solution for forward problem (LAMBDA):

\begin{eqnarray} \label{nove}    
\widehat{I}=I_{exact}\left(1+\sigma\mu\right)
\end{eqnarray}
where $\sigma$ is the noise standard deviation and $\mu$ is a random variable taken from a Gaussian distribution with zero mean and unitary variance. In all simulations we used $\delta=0.05$ and $\delta=0.10$.

Overall, more than fifty pairs of rates of emission of pollutants and their concentrations are necessary for the process of inversion. Similar data sets were used for the stages of activation and the general ANN.

\section{Results}
\label{res}

\begin{figure}[!ht]
	\centering
	\includegraphics[width=3in]{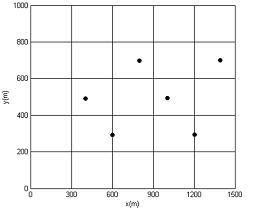}
	\caption{Computational domain divided into 25 subdomains.}
	\label{fig:fig1}
\end{figure}

The area of the numerical experiment used is the same employed by Roberti (2005) \cite{roberti} and Luz (2007) \cite{luz}. The domain is divided into 25 sub-domains, where each cell has the size: 300 m (width) x 200 m (length) - volume of each cell is 60.000.000 $m^{3}$ (height = 1000 m) \cite{roberti}. \autoref{fig:fig1} shows the different subdomains of emissions of contaminants. In this figure, there are sensors, where their positions are represented by $\bullet$ in the area of study.

Six sensors are used inside the domain, and the sensor size is a small volume: 0.1 m x 0.1 m x 0.1 m, positioned at a height of 10 meters, installed in the area, according to Table \autoref{tab:position}.

\autoref{tab:data} shows the meteorological data used by the LAMBDA model to simulate the dispersion of particles, taken from the Copenhagen experiment \cite{roberti, ferrero95}. Meteorological data are speeds and direction averages of wind, measured at three levels, for five different time periods measured on 19/10/1978 \cite{roberti}.

The results obtained with noiseless data are in excellent agreement with the true model. The MLP network produced good estimation of the rate of emission of pollutants compared to the Quasi-Newton method \cite{roberti} and Particle Swarm Optimization (PSO) \cite{luz}. The best results for noisy data were obtained for the MLP network with 15 and 30 neurons in hidden layer and $\delta=0.05$.

\begin{figure}[!ht]
\centering

\subfigure[\label{fig:figure1a.jpg}]
{\includegraphics[width=2in]{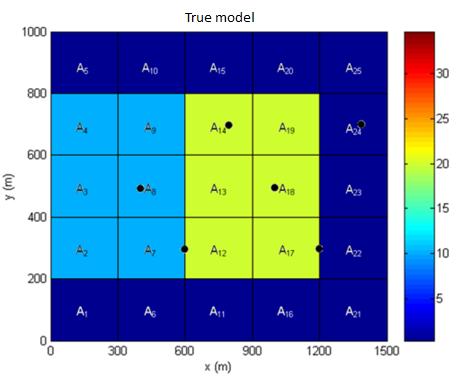}}
\subfigure[\label{fig:figure1b}]
{\includegraphics[width=2in]{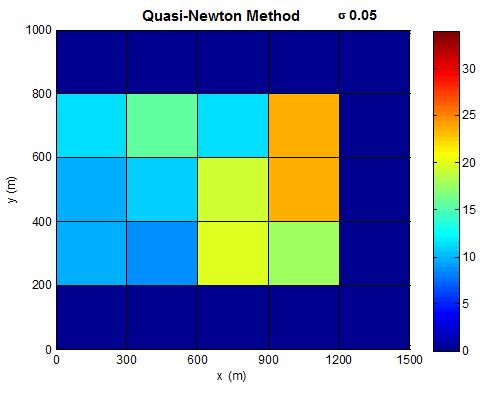}}
\subfigure[\label{fig:figure1c}]
{\includegraphics[width=2in]{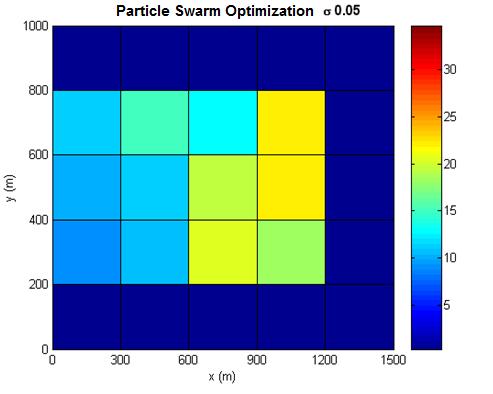}}
\subfigure[\label{fig:figure1d}]
{\includegraphics[width=2in]{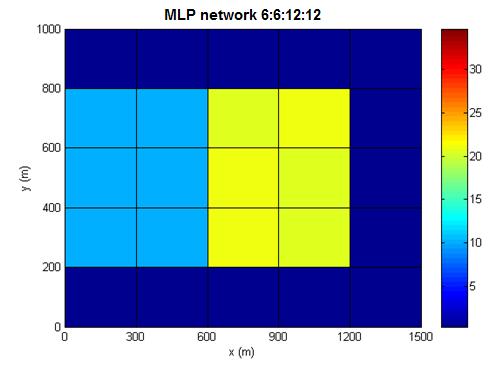}} 
\subfigure[\label{fig:figure1e}]
{\includegraphics[width=2in]{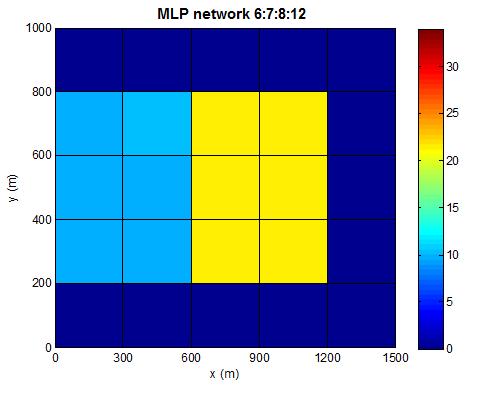}}
\subfigure[\label{fig:figure1f}]
{\includegraphics[width=2in]{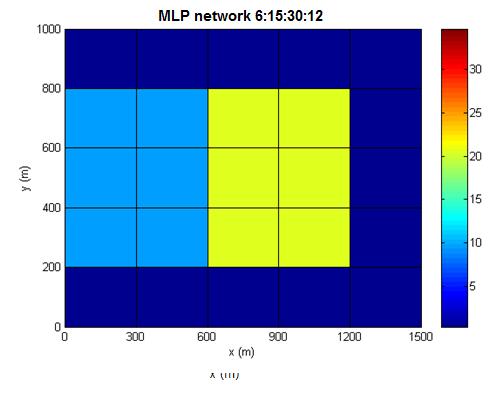}}
\center
\caption{Generalization results of rate of emission of pollutant (in $gm^{-3}s^{-1}$) for noiseless data using $\sigma = 0.05$. (a) True model; (b) Quasi-Newton \cite{roberti}; (c) PSO \cite{luz}; (d)6 and 12 neurons hidden layer; (e) 7 and 8; (f) 15 and 30.}
\label{fig:fig2}
\end{figure}

\begin{figure}[!ht]
\centering

\subfigure[\label{fig:figure2a.jpg}]
{\includegraphics[width=2in]{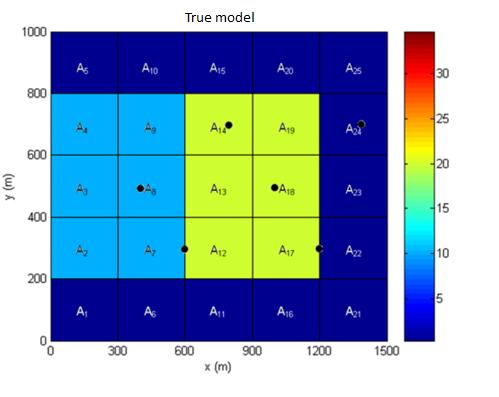}}
\subfigure[\label{fig:figure2b}]
{\includegraphics[width=2in]{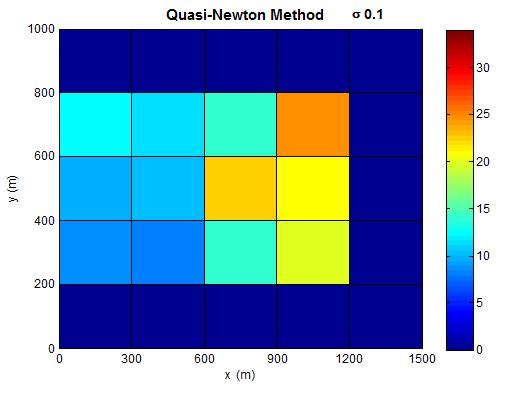}}\\ 
\subfigure[\label{fig:figure2c}]
{\includegraphics[width=2in]{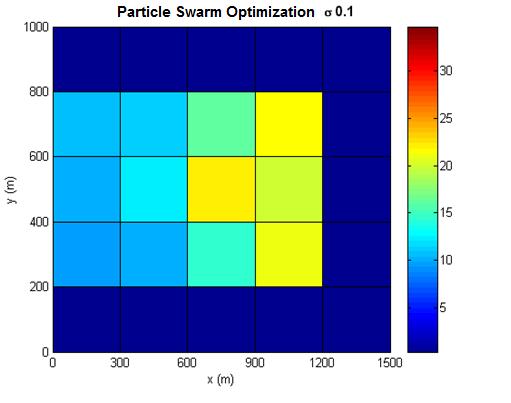}}
\subfigure[\label{fig:figure2d}]
{\includegraphics[width=2in]{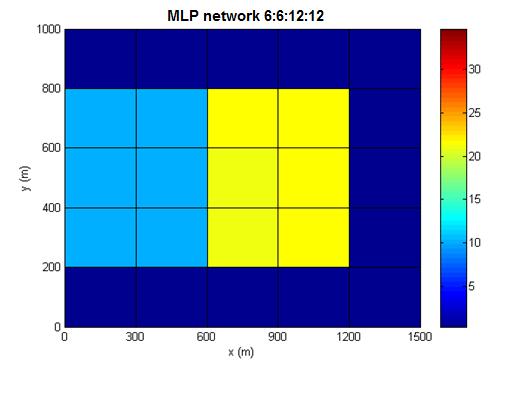}}\\ 
\subfigure[\label{fig:figure2e}]
{\includegraphics[width=2in]{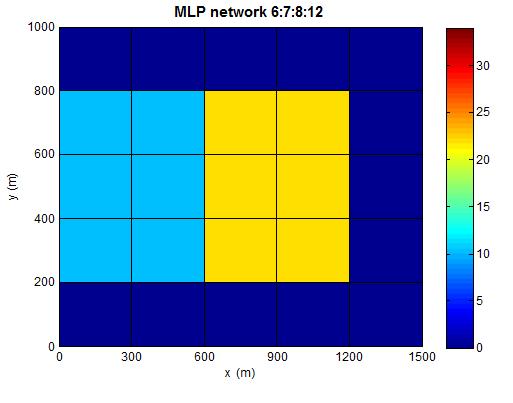}}
\subfigure[\label{fig:figure2f}]
{\includegraphics[width=2in]{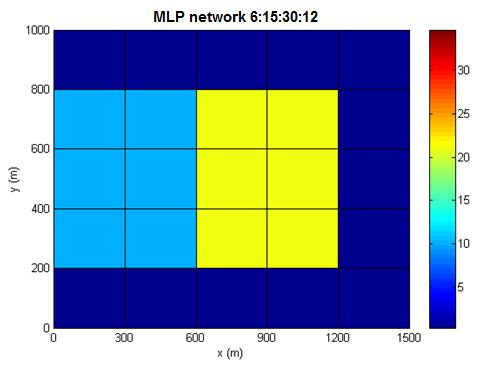}}
\center
\caption{Generalization results of rate of emission of pollutant (in $gm^{-3}s^{-1}$) for noiseless data using $\sigma = 0.10$. (a) True model; (b) Quasi-Newton \cite{roberti}; (c) PSO \cite{luz}; (d)6 and 12 neurons hidden layer; (e) 7 and 8; (f) 15 and 30.}
\label{fig:fig3}
\end{figure}

In order to analyze the performance of the ANNs for the estimation of the rate of gas emission, two experiments were performed. In the first experiment 5\% of white Gaussian noise ($\delta=0.05$) was added to the synthetic experimental data. In the second one, 10\% of noise is used to simulate the real experimental data. All ANNs were trained with two hidden layers, varying the number of hidden neurons and the database training. Several tests were carried out to reach in a good neural network architecture. For all configurations, two hidden layers were considered: (a) ANN-1: 6 and 12, (b) ANN-2: 7 and 8, (c) ANN-3: 15 and 30 neurons in hidden layers. The ANN-3 obtained the best result compared with the model true. \autoref{fig:fig2} and \autoref{fig:fig3}, the topology of the RNA is represented as: $x_{1}: x_{2}: x_{3}: x_{4}$ where $x_{1}$: number of neurons in input layer, $x_{2}$: number of neurons in the 1st hidden layer, $x_{3}$: number of neurons in the 2nd hidden layer and $x_{4}$: Number of neurons in output layer. 

The training phase was carried out until a maximum number of iterations were reached. \autoref{tab:results005} shows the exact results (LAMBDA), the results obtained with regularized inversion (the optimization problem solved by quasi-Newton (Q-N, deterministic) \cite{roberti} and particle swarm optimization (PSO, stochastic) \cite{luz} schemes), and the results obtained with ANN for $\delta=0.05$. The \autoref{tab:results010} shows the same results, but considering $\delta=0.10$.

Figures \autoref{fig:fig2} and \autoref{fig:fig3} shows the results of the generalization tests in comparison with the true model.
\section{Conclusion}
\label{con}

The problem for identifying the minority gas emission rate for the system ground-atmosphere is an important issue for the bio-geochemical cycle, and it has being intensively investigated. This inverse problem has been solved using regularized solutions \cite{seibert}, Bayes estimation \cite{enting, gimson}, and variational methods \cite{Elbern} (this is approach started from the data assimilation studies).

Our previous studies were initiated using generalized least square scheme, with entropic regularization \cite{roberti, roberti2005, luz, luzetal} (for the use of maximum entropy principle on this issue, see also \cite{bocquet, davoine}).

ANNs were used as effective tool for solving the inverse problem of estimation of the rate of gas surface emission. The obtained reconstructions with the MLP showed to be better than the obtained with regularization methods \cite{roberti, luz}. Another advantage of the use of neural network is, after the training phase, the reconstruction algorithm is faster than the regularized inversion methods. Neural inversion is unique scheme that does not need a solution of the associated forward problem.

In practice, operational inversion algorithms reduce the risk of being trapped in local minima by starting the iterative search process from an initial guess solution that is sufficiently close to the true profile. However, the dependence of the final solution on a good choice of the initial guess represents a fundamental weakness of such algorithms, particularly in regions where less a priori information is available \cite{kukkonen}. ANN approaches can relax this constraint incorporating more data in the dataset during the learning phase.

The ANNs can be inaccurate if they are used to extrapolate to cases outside the training domain. However the use of ANN techniques can provide good solutions when the training phase encompasses the domain of the potential solutions to the real problem.

\begin{table}[!ht]
	\centering
	\scriptsize
    \setlength{\tabcolsep}{9pt}
	\caption{Position of the sensors in the area.}
\begin{tabular}{|l|l|l|}
\hline\multicolumn{1}{c|}{Sensor} & \multicolumn{1}{c|}{Position x (m)} & \multicolumn{1}{c}{Position y (m)} \\ 
\hline
\multicolumn{1}{c|}{1} & \multicolumn{1}{c|}{400} & \multicolumn{1}{c}{500} \\ 
\hline
\multicolumn{1}{c|}{2} & \multicolumn{1}{c|}{600} & \multicolumn{1}{c}{300} \\ 
\hline
\multicolumn{1}{c|}{3} & \multicolumn{1}{c|}{800} & \multicolumn{1}{c}{700} \\ 
\hline
\multicolumn{1}{c|}{4} & \multicolumn{1}{c|}{1000} & \multicolumn{1}{c}{500} \\ 
\hline
\multicolumn{1}{c|}{5} & \multicolumn{1}{c|}{1200} & \multicolumn{1}{c}{300} \\ 
\hline
\multicolumn{1}{c|}{6} & \multicolumn{1}{c|}{1400} & \multicolumn{1}{c}{700} \\ 
\hline
\end{tabular}
\label{tab:position}
\end{table}

\begin{table}[!ht]
	\centering
	\scriptsize
    \setlength{\tabcolsep}{9pt}
	\caption{Meteorological data used in the experiment Copenhagen.}
\begin{tabular}{l|l|l|l|l|l|l}
\hline
Time & \multicolumn{3}{c|}{Speed $\overline{U}$ (m/s)} & \multicolumn{3}{c}{Direction $\overline{U}$ (º)} \\ 
(h:m) & \multicolumn{1}{l}{10 m} & \multicolumn{1}{l}{120 m} & 200 m  & \multicolumn{1}{l}{10 m } & \multicolumn{1}{l}{120 m} & 200 m \\ 
\hline
12:05 & 2,6 & 5,7 & 5,7 & 290 & 310 & 310 \\ 
\hline
12:15 & 2,6 & 5,1 & 5,7 & 300 & 310 & 310 \\ 
\hline
12:25 & 2,1 & 4,6 & 5,1 & 280 & 310 & 320 \\ 
\hline
12:35 & 2,1 & 4,6 & 5,1 & 280 & 310 & 320 \\ 
\hline
12:45 & 2,6 & 5,1 & 5,7 & 290 & 310 & 310 \\ 
\hline
\end{tabular}
\label{tab:data}
\end{table}

\begin{table}[!ht]
\raggedleft
\centering
 \scriptsize
    \setlength{\tabcolsep}{8pt}

	\caption{Results of estimation of the rate o gas emission ($gm^{-3}s^{-1}$), for noise level $\sigma = 0.05$.}
\begin{tabular}{l|l|l|l|l|l}
\hline
\multicolumn{1}{c|}{cell} & \multicolumn{1}{c|}{Exact} & \multicolumn{1}{c|}{Q-N ($gm^{-3}s^{-1}$)} & \multicolumn{1}{c|}{PSO ($gm^{-3}s^{-1}$)} & \multicolumn{1}{c|}{ANN 6:6:12:12} & \multicolumn{1}{c}{ANN 6:15:30:12} \\ 
\multicolumn{1}{c|}{} & \multicolumn{1}{c|}{($gm^{-3}s^{-1}$)} & \multicolumn{1}{c|}{(Roberti, 2005)} & \multicolumn{1}{c|}{(Luz, 2007)} & \multicolumn{1}{c|}{($gm^{-3}s^{-1}$)} & \multicolumn{1}{c}{($gm^{-3}s^{-1}$)} \\ 
\hline
\multicolumn{1}{c|}{$A_{2}$} & \multicolumn{1}{c|}{10} & \multicolumn{1}{c|}{9,82} & \multicolumn{1}{c|}{09,34} & \multicolumn{1}{c|}{09,92} & \multicolumn{1}{c}{09,79} \\ 
\hline
\multicolumn{1}{c|}{$A_{3}$} & \multicolumn{1}{c|}{10} & \multicolumn{1}{c|}{9,63} & \multicolumn{1}{c|}{10,07} & \multicolumn{1}{c|}{09,83} & \multicolumn{1}{c}{09,74} \\ 
\hline
\multicolumn{1}{c|}{$A_{4}$} & \multicolumn{1}{c|}{10} & \multicolumn{1}{c|}{11,26} & \multicolumn{1}{c|}{11,26} & \multicolumn{1}{c|}{09,86} & \multicolumn{1}{c}{04,81} \\ 
\hline
\multicolumn{1}{c|}{$A_{7}$} & \multicolumn{1}{c|}{10} & \multicolumn{1}{c|}{8,76} & \multicolumn{1}{c|}{10,95} & \multicolumn{1}{c|}{09,82} & \multicolumn{1}{c}{09,80} \\ 
\hline
\multicolumn{1}{c|}{$A_{8}$} & \multicolumn{1}{c|}{10} & \multicolumn{1}{c|}{11,06} & \multicolumn{1}{c|}{10,93} & \multicolumn{1}{c|}{09,71} & \multicolumn{1}{c}{09,73} \\ 
\hline
\multicolumn{1}{c|}{$A_{9}$} & \multicolumn{1}{c|}{10} & \multicolumn{1}{c|}{15,51} & \multicolumn{1}{c|}{14,99} & \multicolumn{1}{c|}{09,71} & \multicolumn{1}{c}{09,79} \\ 
\hline
\multicolumn{1}{c|}{$A_{12}$} & \multicolumn{1}{c|}{20} & \multicolumn{1}{c|}{20,12} & \multicolumn{1}{c|}{20,79} & \multicolumn{1}{c|}{20,94} & \multicolumn{1}{c}{20,61} \\ 
\hline
\multicolumn{1}{c|}{$A_{13}$} & \multicolumn{1}{c|}{20} & \multicolumn{1}{c|}{19,25} & \multicolumn{1}{c|}{19,83} & \multicolumn{1}{c|}{20,81} & \multicolumn{1}{c}{20,61} \\ 
\hline
\multicolumn{1}{c|}{$A_{14}$} & \multicolumn{1}{c|}{20} & \multicolumn{1}{c|}{11,52} & \multicolumn{1}{c|}{13,06} & \multicolumn{1}{c|}{20,93} & \multicolumn{1}{c}{20,60} \\ 
\hline
\multicolumn{1}{c|}{$A_{17}$} & \multicolumn{1}{c|}{20} & \multicolumn{1}{c|}{17,88} & \multicolumn{1}{c|}{18,72} & \multicolumn{1}{c|}{20,94} & \multicolumn{1}{c}{20,60} \\ 
\hline
\multicolumn{1}{c|}{$A_{18}$} & \multicolumn{1}{c|}{20} & \multicolumn{1}{c|}{23,82} & \multicolumn{1}{c|}{22,76} & \multicolumn{1}{c|}{20,96} & \multicolumn{1}{c}{20,61} \\ 
\hline
\multicolumn{1}{c|}{$A_{19}$} & \multicolumn{1}{c|}{20} & \multicolumn{1}{c|}{23,44} & \multicolumn{1}{c|}{22,47} & \multicolumn{1}{c|}{20,77} & \multicolumn{1}{c}{20,59} \\ 
\hline
\end{tabular}
\label{tab:results005}
\end{table}

\begin{table}[!ht]
	\centering
	\scriptsize
    \setlength{\tabcolsep}{8pt}
    
	\caption{Results of estimation of the rate o gas emission ($gm^{-3}s^{-1}$), for noise level $\sigma = 0.10$.}
\begin{tabular}{l|l|l|l|l|l}
\hline
\multicolumn{1}{c|}{cell} & \multicolumn{1}{c|}{Exact} & \multicolumn{1}{c|}{Q-N ($gm^{-3}s^{-1}$)} & \multicolumn{1}{c|}{PSO ($gm^{-3}s^{-1}$)} & \multicolumn{1}{c|}{ANN 6:6:12:12} & \multicolumn{1}{c}{ANN 6:15:30:12} \\ 
\multicolumn{1}{c|}{} & \multicolumn{1}{c|}{($gm^{-3}s^{-1}$)} & \multicolumn{1}{c|}{(Roberti, 2005)} & \multicolumn{1}{c|}{(Luz, 2007)} & \multicolumn{1}{c|}{($gm^{-3}s^{-1}$)} & \multicolumn{1}{c}{($gm^{-3}s^{-1}$)} \\ 
\hline
\multicolumn{1}{c|}{$A_{2}$} & \multicolumn{1}{c|}{10} & \multicolumn{1}{c|}{8,97} & \multicolumn{1}{c|}{09,83} & \multicolumn{1}{c|}{10,33} & \multicolumn{1}{c}{10,11} \\ 
\hline
\multicolumn{1}{c|}{$A_{3}$} & \multicolumn{1}{c|}{10} & \multicolumn{1}{c|}{9,97} & \multicolumn{1}{c|}{10,40} & \multicolumn{1}{c|}{10,11} & \multicolumn{1}{c}{10,11} \\ 
\hline
\multicolumn{1}{c|}{$A_{4}$} & \multicolumn{1}{c|}{10} & \multicolumn{1}{c|}{12,52} & \multicolumn{1}{c|}{10,79} & \multicolumn{1}{c|}{10,12} & \multicolumn{1}{c}{10,20} \\ 
\hline
\multicolumn{1}{c|}{$A_{7}$} & \multicolumn{1}{c|}{10} & \multicolumn{1}{c|}{7,98} & \multicolumn{1}{c|}{10,50} & \multicolumn{1}{c|}{10,27} & \multicolumn{1}{c}{10,20} \\ 
\hline
\multicolumn{1}{c|}{$A_{8}$} & \multicolumn{1}{c|}{10} & \multicolumn{1}{c|}{10,14} & \multicolumn{1}{c|}{12,06} & \multicolumn{1}{c|}{10,24} & \multicolumn{1}{c}{10,06} \\ 
\hline
\multicolumn{1}{c|}{$A_{9}$} & \multicolumn{1}{c|}{10} & \multicolumn{1}{c|}{11,56} & \multicolumn{1}{c|}{11,28} & \multicolumn{1}{c|}{10,17} & \multicolumn{1}{c}{10,17} \\ 
\hline
\multicolumn{1}{c|}{$A_{12}$} & \multicolumn{1}{c|}{20} & \multicolumn{1}{c|}{13,84} & \multicolumn{1}{c|}{14,56} & \multicolumn{1}{c|}{21,21} & \multicolumn{1}{c}{20,97} \\ 
\hline
\multicolumn{1}{c|}{$A_{13}$} & \multicolumn{1}{c|}{20} & \multicolumn{1}{c|}{22,65} & \multicolumn{1}{c|}{22,67} & \multicolumn{1}{c|}{21,28} & \multicolumn{1}{c}{21,00} \\ 
\hline
\multicolumn{1}{c|}{$A_{14}$} & \multicolumn{1}{c|}{20} & \multicolumn{1}{c|}{14,14} & \multicolumn{1}{c|}{15,85} & \multicolumn{1}{c|}{21,32} & \multicolumn{1}{c}{20,95} \\ 
\hline
\multicolumn{1}{c|}{$A_{17}$} & \multicolumn{1}{c|}{20} & \multicolumn{1}{c|}{19,99} & \multicolumn{1}{c|}{21,56} & \multicolumn{1}{c|}{21,47} & \multicolumn{1}{c}{20,99} \\ 
\hline
\multicolumn{1}{c|}{$A_{18}$} & \multicolumn{1}{c|}{20} & \multicolumn{1}{c|}{21,17} & \multicolumn{1}{c|}{20,05} & \multicolumn{1}{c|}{21,47} & \multicolumn{1}{c}{20,99} \\ 
\hline
\multicolumn{1}{c|}{$A_{19}$} & \multicolumn{1}{c|}{20} & \multicolumn{1}{c|}{24,90} & \multicolumn{1}{c|}{21,74} & \multicolumn{1}{c|}{21,48} & \multicolumn{1}{c}{20,96} \\ 
\hline
\end{tabular}
\label{tab:results010}
\end{table}

\end{document}